\documentclass[runningheads]{llncs}
\usepackage{graphicx}
\usepackage{hyperref}
\begin{document}
\title{Understanding Bias in Machine Learning}
%
%
\author{Jindong Gu$^{1,2}$,   Daniela Oelke$^2$ \\}
\authorrunning{}
%
\institute{$^1$The University of Munich\\
$^2$Siemens AG, Corporate Technology\\}
\maketitle              
\begin{abstract}
Bias is known to be an impediment to fair decisions in many domains such as human resources, the public sector, health care etc. Recently, hope has been expressed that the use of machine learning methods for taking such decisions would diminish or even resolve the problem. At the same time, machine learning experts warn that machine learning models can be biased as well.

In this article, our goal is to explain the issue of bias in machine learning from a technical perspective and to illustrate the impact that biased data can have on a machine learning model. To reach such a goal, we develop interactive plots to visualizing the bias learned from synthetic data. The interactive plots are available \footnote{please visit the \href{https://visxai.io/2018.html}{homepage} of the 1st Workshop on Visualization for AI Explainability \\ or run our \href{https://github.com/Jindong-Explainable-AI/Bias_in_Machine_Learning}{python code}.}.

\end{abstract}
\section{How does bias get into a machine learning model?}
To be able to let the machine take a decision automatically, we have to teach it how to do it right. One way to do so is to formulate explicitly as a rule when to take which decision. However, many situations are too complex for this. So what can we do? A central idea of machine learning is that the machine learns the rules and patterns by itself from examples. The examples are decisions that humans have taken in the past together with the information about the subject (the data) that they based their decision upon. We call this kind of data “training data”, because the machine uses it to learn to take a decision as a human would have done it.

Now it should be clear, why a machine learning model can be biased as well: If the data or the decisions taken on it are biased and the machine uses them as an example, then the machine is going to incorporate this bias into the model. It learns the bias from the examples given to it.

So the first thing that you should keep in mind is this: Bias gets into the model through the data that is used for building the machine learning model.

\section{What types of bias exist?}
Machine learning experts distinguish three types of bias. Thereby, the term “bias” does not only refer to bias that leads to discriminating or unfair decisions. Generally speaking, data is biased if the sampling distribution (that means the data which we use for training the model) is different from the population distribution (referring to the true situation in the real world). Putting it the other way round, to avoid bias we have to make sure that the data sample that we use for training the model resembles as closely as possible the true distribution of the features and the decisions taken on them.

Before we explain the three types of bias, let us first define what we mean by training data. We already know that it can be considered as a set of examples. Each example consists of two parts: a) the correct decision, also called a target or label and b) the attributes describing the example, also called the features.

Let’s make this more concrete. Imagine that our goal would be to sort a stack of applications into good candidates for a job (those that a recruiter should invite to a job interview and those that do not fit the requirements and can be rejected right away). In this case the features could be specific qualifications that an applicant may or may not have, the university that (s)he got her/his degree from, the age, the gender etc. The label would be the decision that a recruiter has made based on this information about the applicant. We would use data from the past when the decision still was taken manually to train the model. Our goal is to be able to classify an applicant as fit or no-fit only based on the features, i.e. the information that we are provided with in the application as a recruiter would do.

The following three types of bias might occur in such a data set:

1) \textbf{Covariate Shift} occurs if one of the features is not covered uniformly in the dataset. In our example, there may be one skill for which we only have examples of candidates with little experience in the specific area. A reason for this might be that the skill was not taught at universities until a couple of years ago (e.g. a new programming language). 

2) \textbf{Sample Selection Bias} refers to a correlation between a (subset of) feature(s) and the label. If this correlation only occurs in our set of examples but not in the normal population our dataset is biased. If in the past, applicants of a specific race were systematically discriminated, then this feature would correlate with the fit / no-fit label that we want to predict. Another example might be that a certain combination of skills has only been observed for applicants coming from a specific university in the past because only this university offered a field of study in which both skills are relevant. Then the university feature would be correlated to our label although in general we would not say that a candidate is suitable for the job only because (s)he graduated from a specific university.

3) \textbf{Imbalance Bias} denotes the situation in which there are considerably fewer examples for one specific decision (label) than for the other(s). In our example, there may be considerably more people that were rejected in the past than examples for applicants that finally were accepted.

Note that in reality these types of bias do not necessarily occur separately, but often a biased dataset contains a mixture of these. Furthermore, the examples illustrate that not every bias necessarily results from unconscious bias of humans or a discrimination of certain groups of people. Sometimes bias occurs naturally in the data and it may or may not affect the decisions that a machine learning model which was trained on it takes. We will take a closer look at this in the following sections.

As a side note: In the machine learning community, the term “bias” also refers to the difference between an estimator’s expected value and the true value of the estimated variable. In this article though, we focus on its definition as the difference between the sampling distribution and the population distribution.

\section{Exploring the impact of bias}
Let’s take a simplified IT-department job application scenario to illustrate the impact of biased data on a neural network. We assume that the human resources department (HR) of our fictional company bases its reject / invite decisions only on four characteristics of applicants, namely, the statistical knowledge, the Python programming skills, the Pytorch programming skills and the Matlab programming skills. Based on a test, each applicant gets a score between 0 and 1 (0: no experience, 1: expert) for each of these four skills. In addition, the university they graduated from is known for all applicants. The HR department decides to invite all applicants to an interview whose score is higher than 0.7 in any two skills, regardless of the university they come from.  

A couple of years later, the HR department decides to automatize this pre-selection of candidates.  A data analyst is given the scores of the applicants of the past together with the final decisions of the HR department and is asked to train a machine learning model on this dataset. 

This is of course an oversimplification of the complex job recruiting processes of a company. But by keeping the data and the selection criterion simple and clear, the effect that bias in the data may have on a machine learning model can be better illustrated.

In the following, we will discuss what kind of bias such a dataset may have. We will show the impact of the bias on the resulting machine learning model and visualize the decisions of the model. By interacting with the visualized results, you can get a better understanding of the three types of bias. To make the difference clear, we will focus on each type of bias separately, and will remove the other two types of bias from the dataset before we do our tests.

\subsection{Covariate Shift}
Pytorch is a relatively new deep learning framework. Because of this, a couple of years ago when the data was collected, there was no applicant that scored above 0.7. This can easily be seen if we visualize the data in a Parallel Coordinates Plot. Before we continue, we would like to explain you how to read the visualization and how you can interact with it.

In a Parallel Coordinates Plot (PC Plot) each feature of our dataset is represented as a vertical axis. A data record (in this case representing an applicant) is added to the plot by drawing a line from axis to axis in a way that it cuts each axis at the level of the score of the applicant for the specific feature.

You can interact with the visualization in several ways (please visit our web-based version):
\begin{enumerate}
\item It is possible to change the order of the axes by dragging the axis’ label to another position. Because only relationships between adjacent axes can be seen, changing the order of the axes can reveal new correlations that so far were not visible.
\item You can select / brush a range of values on each axis by dragging the mouse around it. Lines that run through ALL selected ranges on all axes are highlighted in pink. This way the relationship between features can be explored across multiple features.
\item To clear a selection, simply click on the pink bar which marks the selected range on an axis. 
\end{enumerate}

See \cite{inselberg1990parallel} for more information about parallel coordinates.
\begin{figure}[h]
        \centering
        \includegraphics[width=\textwidth]{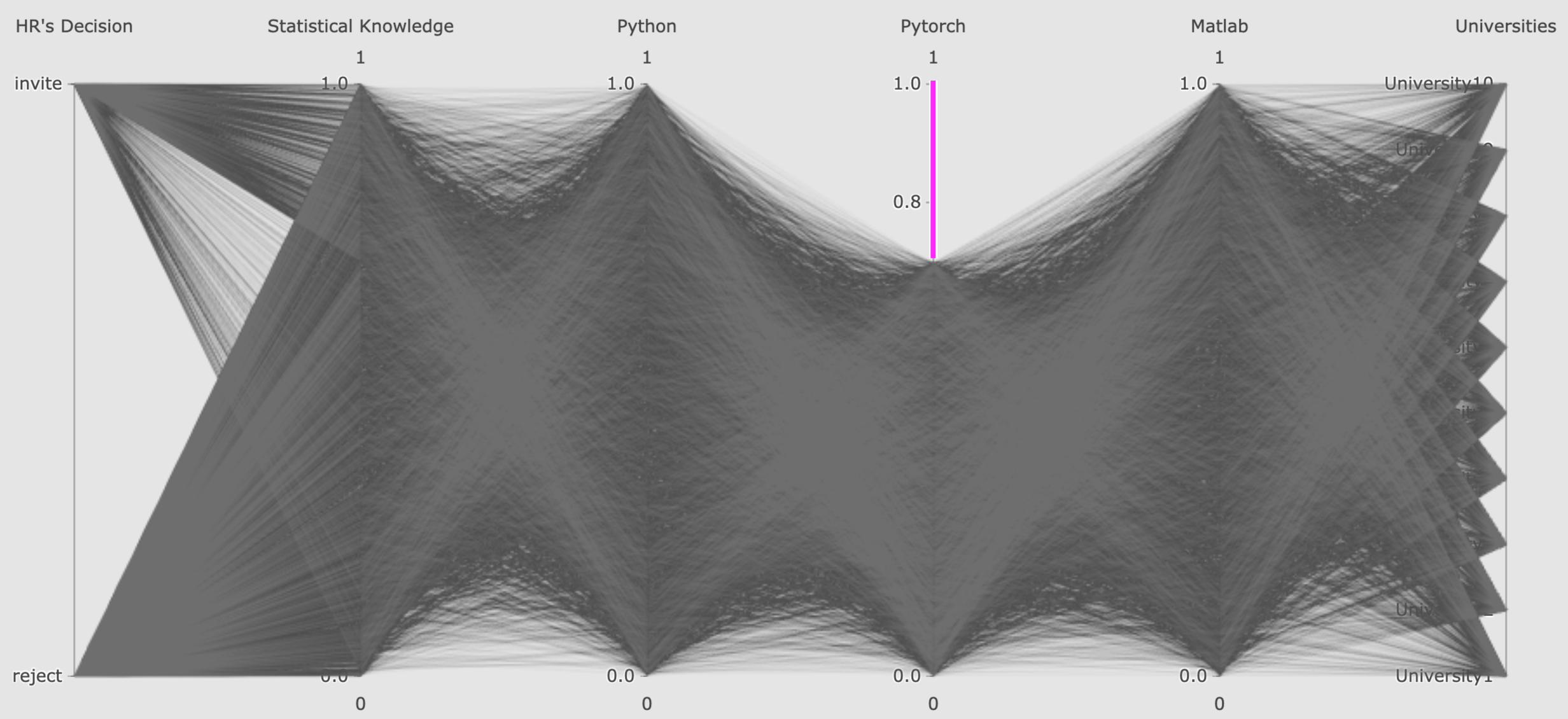}
        \caption{Interactive Visualization of Data about Previous Applicants}
        \label{fig:CovarS}
\end{figure}
In Figure \ref{fig:CovarS}, as you can see in the PC plot that visualizes the training data, there are no applicants whose Pytorch score is higher than 0.7.
\begin{figure}[h]
        \centering
        \includegraphics[width=\textwidth]{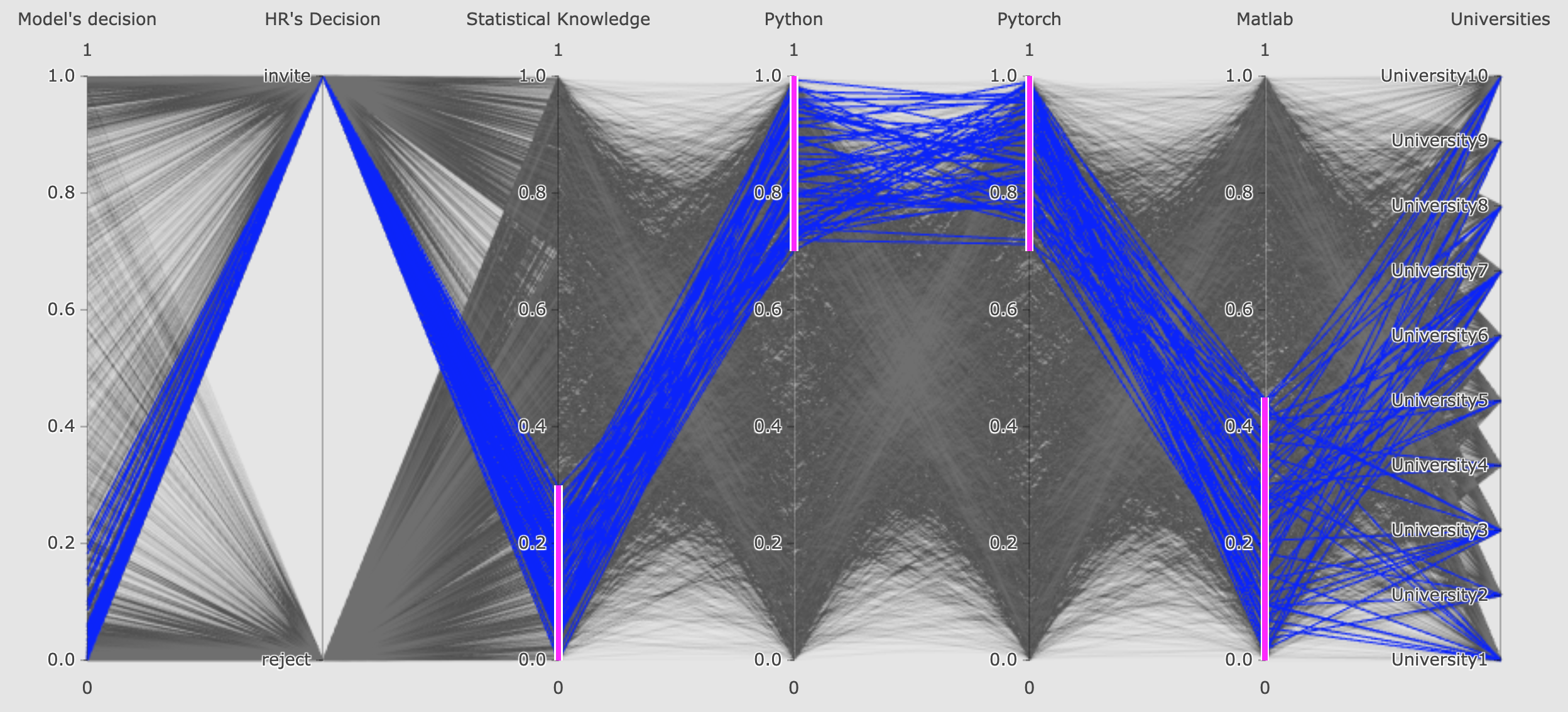}
        \caption{Interactive Visualization of Data about New Coming Applicants}
        \label{fig:CovarS_test}
\end{figure}

This looks different if we visualize the scores of today’s applicants in Figure \ref{fig:CovarS_test}. Now, there are quite a lot of candidates for the job that have a solid knowledge in Pytorch. We can conclude that there is clearly a Covariate Shift bias in our data.

But how did the Covariate Shift now affect the results of our machine learning model? To be able to assess this we need test data. Test data is data for which we know the true label (in this case “reject” or “invite”). This so-called “ground-truth” allows us to a compare the predicted labels of the model to the decisions that an employee of the HR department would have made. The respective axes in the plot are called “HR’s decision” (either an invite or reject) and “Model’s decision” (a probability value between 0 and 1 with 0 denoting a clear reject recommendation and 1 a clear recommendation to invite the candidate).

If you brush the range 0.7 to 1 on the Python and the Pytorch axis and select a range of lower values for the two other skills, you can see that the machine learning model is strongly in favor of rejecting most of these applicants although these applicants would satisfy our criterion of having high scores in at least two skills. The reason for this is that the model has not seen any examples for applicants with high Pytorch skills in the training data (and therefore also has not seen any examples for applicants with a high Pytorch skill that got invited). As a consequence, its predictions are unreliable for such candidates. If we instead select applicants which score high in two other skills (excluding Pytorch), the plot shows that the likelihood that our model suggests to invite them is much higher.

[As a side note: All applicants in our training data which had high scores (above 0.7) in two skills were invited to a job interview by the HR department. The PC plot of the test data reveals that the Neural Network did not pick up this pattern but gives low scores to some applicants that scored high in two skills. Still, the induced bias is clearly visible in the plot.]

There is a second type of analysis that we can do to explore the impact of biased data on the machine learning model. We are going to use Layer-wise Relevance Propagation (LRP) which belongs to the tool set of Explainable Artificial Intelligence. Technically speaking, LRP identifies the importance of each input feature for an individual decision by running a backward pass in the neural network. (See \cite{LRP,bach2015pixel} for a more detailed explanation of LRP.) The technique can be used to get an idea what a neural network’s learned model based its decision on for a specific instance.

\begin{figure}[h]
        \centering
        \includegraphics[width=0.9\textwidth]{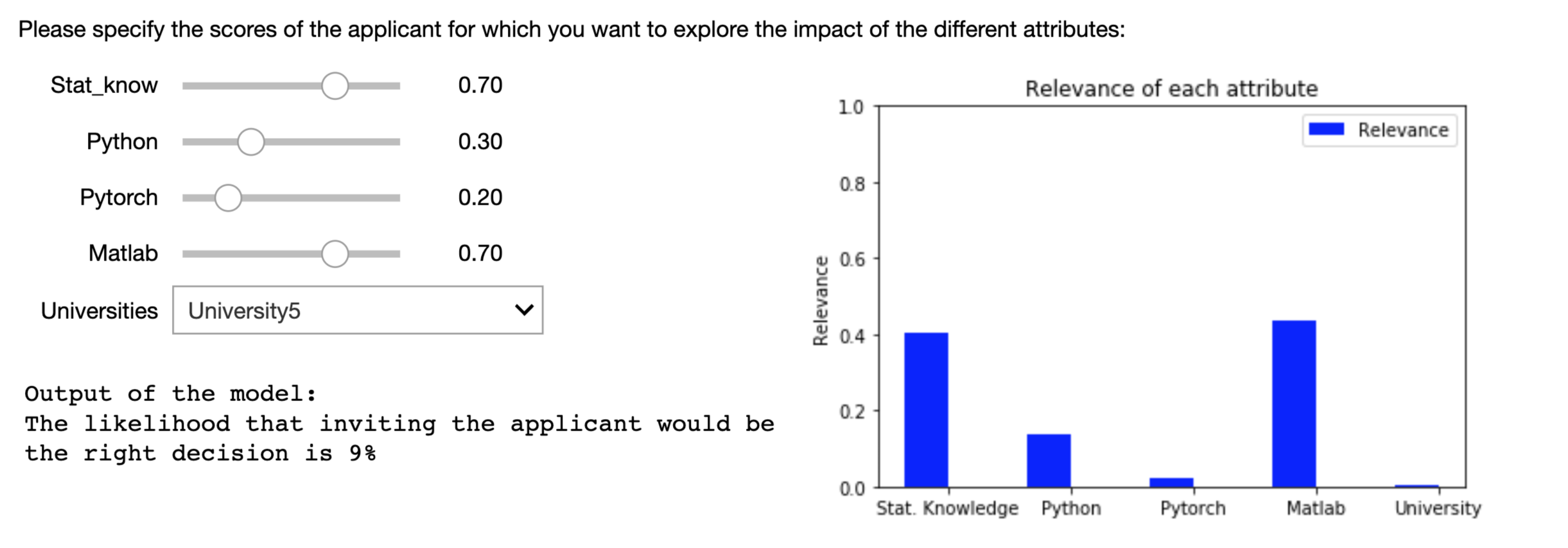}
        \caption{Interactive Visualization of Relevance of each Feature}
        \label{fig:CovarS_inf}
\end{figure}

The interface below (in Figure \ref{fig:CovarS_inf}) allows you to specify the scores of an applicant for which you want to get an explanation. The bar chart next to it shows relevance scores for each feature. The higher the bars the more relevant was the specific feature for the decision of the machine learning model for this specific instance. Note that we use relative relevance here, meaning that a value of 66\% for one feature and 33\% of another feature has to be interpreted as the first feature being twice as relevant for the decision as the second. Below the sliders the likelihood that inviting the applicant would be the right decision (according to the model) is given.

If you play around with the values, you will recognize that features with higher score in general are considered more relevant by the model than the ones with lower scores. However, this is not true for Pytorch. Changing its score has far less effect on its relevance for the decision than for all the other skills. Interestingly, if the score of Python is the same as for one of the other features, the Python feature is considered significantly more relevant for the decision than the other features. Changing the university of the applicant on the other hand has almost no effect. We can conclude that this feature is not considered as important by the model for the decision.

If we only set the scores of two skills to high values and one of them is Pytorch, we can observe the same effect that we have already seen in the Parallel Coordinates Plot: the model is more likely to reject than to accept the applicant and does not consider the Pytorch skill as an important factor in its decision.

\subsection{Sample Selection Bias}
In case of the Sample Selection Bias a correlation between a (subset of) feature(s) and the label exists which only occurs in the training data but not in the normal population. For our scenario let us assume that until a couple of years ago there were only few universities that offered a program which imparted both, a solid knowledge in statistics and at the same time programming skills in python. However, these were exactly the two skills that were most important for the IT department at that time which is why candidates with such a profile were preferred over others (although they were not the only ones that got invited to an interview).

\begin{figure}[h]
        \centering
        \includegraphics[width=\textwidth]{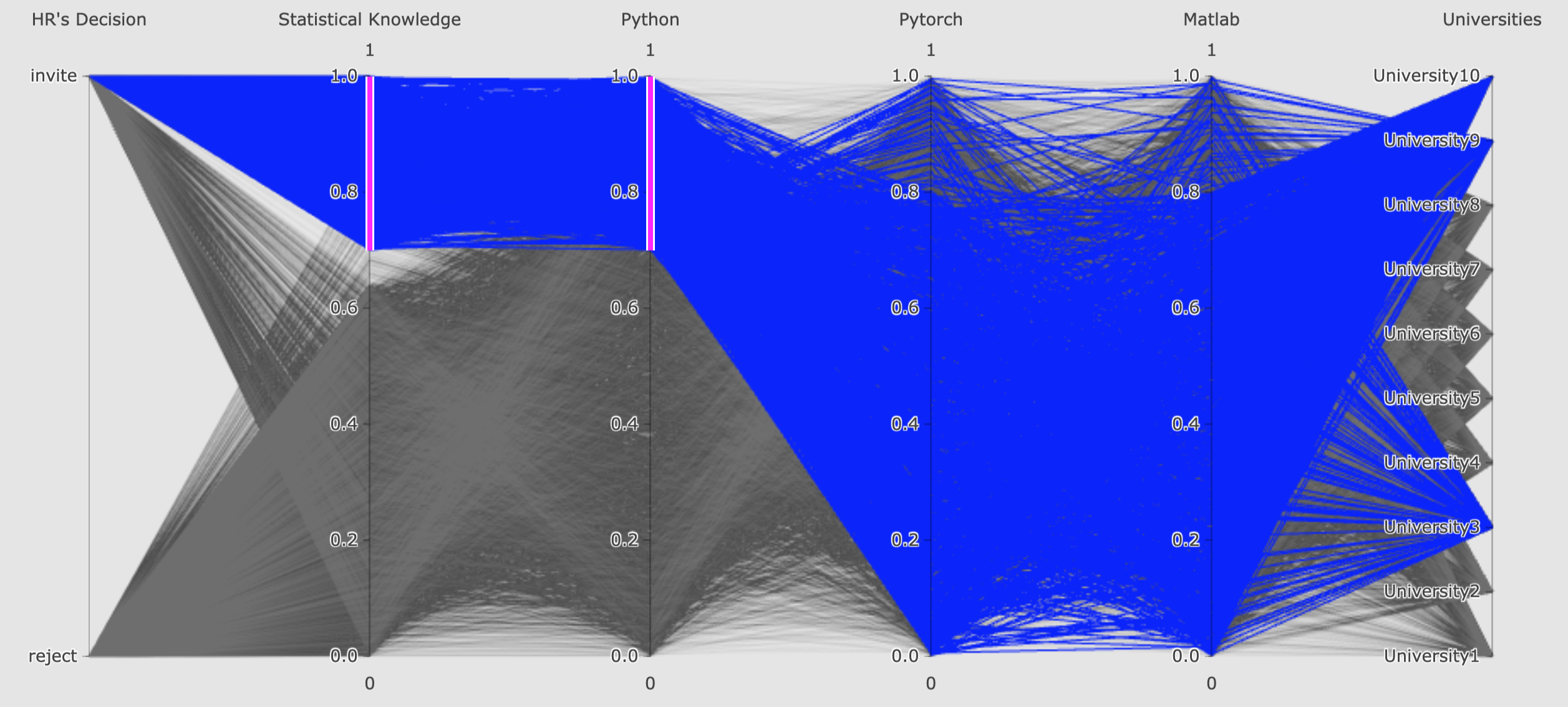}
        \caption{Interactive Visualization of Training Data about Previous Applicants}
        \label{fig:SampleS}
\end{figure}

This becomes apparent when visualizing the training data in Figure \ref{fig:SampleS}. If we brush the 0.7 to 1 range of the axes “Statistical Knowledge” and “Python” most of the applicants are coming from University10. Two additional universities from which at least some candidates have proven expertise in both areas – Statistical Knowledge and Python – are University9 and University3.

As Data Science becomes more and more popular, more universities start to offer similar programs. Will our machine learning model be able to understand that what is important is the skills and not the university the applicant comes from? There are now two aspects in the training data that increase the likelihood that someone has been invited: 1. The ones that got invited had high scores in Statistical knowledge AND Python. 2. They came from one of three different universities with programs that taught the two skills. A machine does not know that semantically only taking the skills into account is meaningful and that the university should not be a relevant factor for the decision.

\begin{figure}[h]
        \centering
        \includegraphics[width=\textwidth]{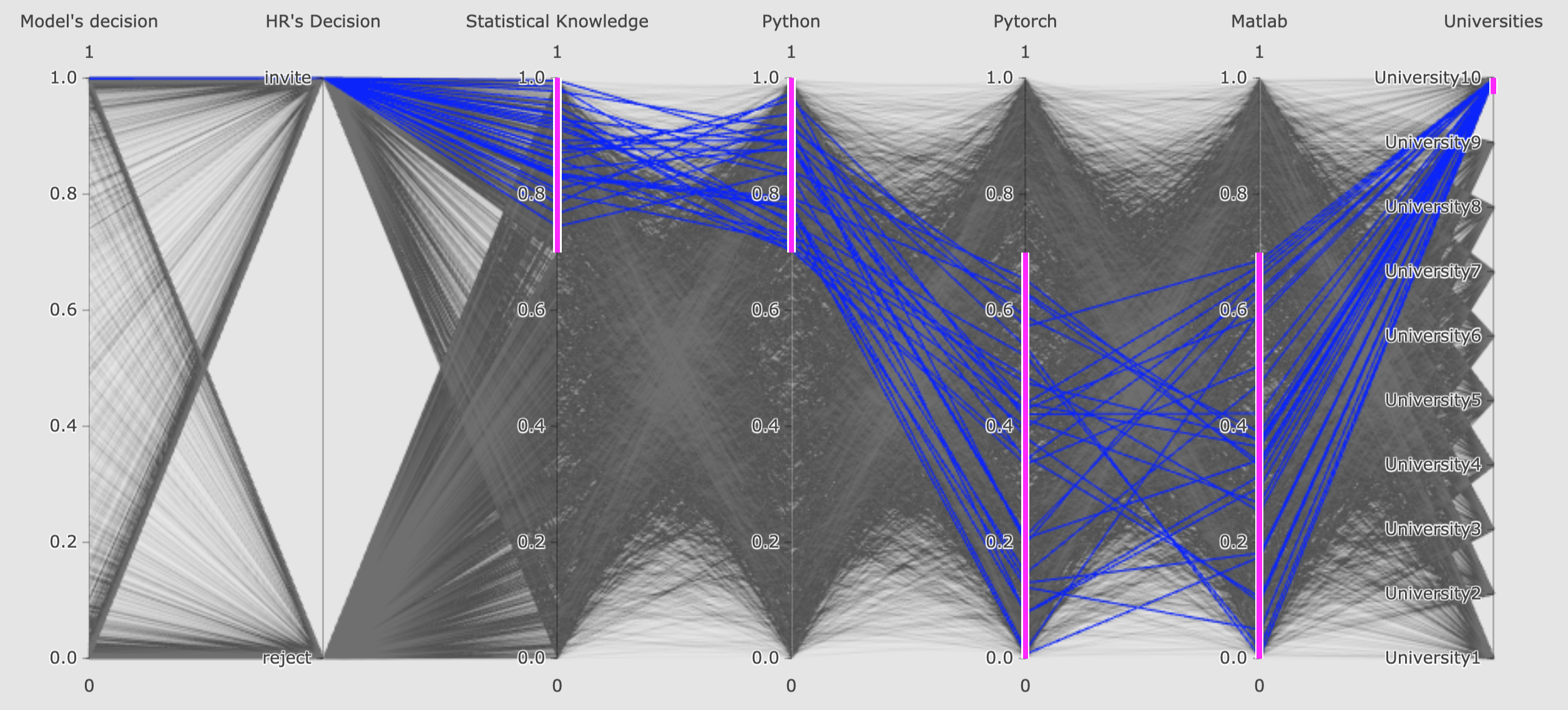}
        \caption{Interactive Visualization of Test Data about New Coming Applicants}
        \label{fig:SampleS_test}
\end{figure}

To explore the model’s behavior we will now use test data again for which we already know the correct decision as the HR department would have taken it (see Figure \ref{fig:SampleS_test}). If you select the 0.7 to 1 range on both the “Statistical Knowledge” axis and the “Python” axis in the plot below and at the same time select lower value ranges for “Pytorch” and “Matlab”, you can see that the model suggests to invite some but not all of these candidates to an interview. To explore the impact that the university the applicant is coming from had on the decision let us now additionally brush the “Universities” axis in a way that only one university is selected. If this is University10, the model outputs a clear “invite” recommendation. Similarly, many applicants from University9 and University3 get such a recommendation. The opposite is the case for applicants from other universities. Here the model tentatively suggests rejecting them. (Hint: Drag the pink range selection bar on the “Universities” axis up and down to explore the decisions made for applicants from the different universities one-by-one.)

You can continue to explore the model by making different selections. How does the model behave if in addition to the “Statistical Knowledge” and “Python” the applicant has high scores in a third skill? Is the decision still influenced by the University the applicant comes from? What characteristics did the applicants have that the model suggested to invite but that did not have high scores for “Statistical Knowledge” and “Python”?

A visualization like this enables gaining an intuition for the model and an increased understanding of the data. Even with an artificial dataset with strong patterns as we use it in our demos, what the neural network learns from the data might not follow the same decision path that a human would take. Therefore, the exploration of the model can reveal interesting patterns in the data that were previously unknown. However, it can also reveal an undesired bias in the data that we have to take care of to get reliable results.

\begin{figure}[h]
        \centering
        \includegraphics[width=\textwidth]{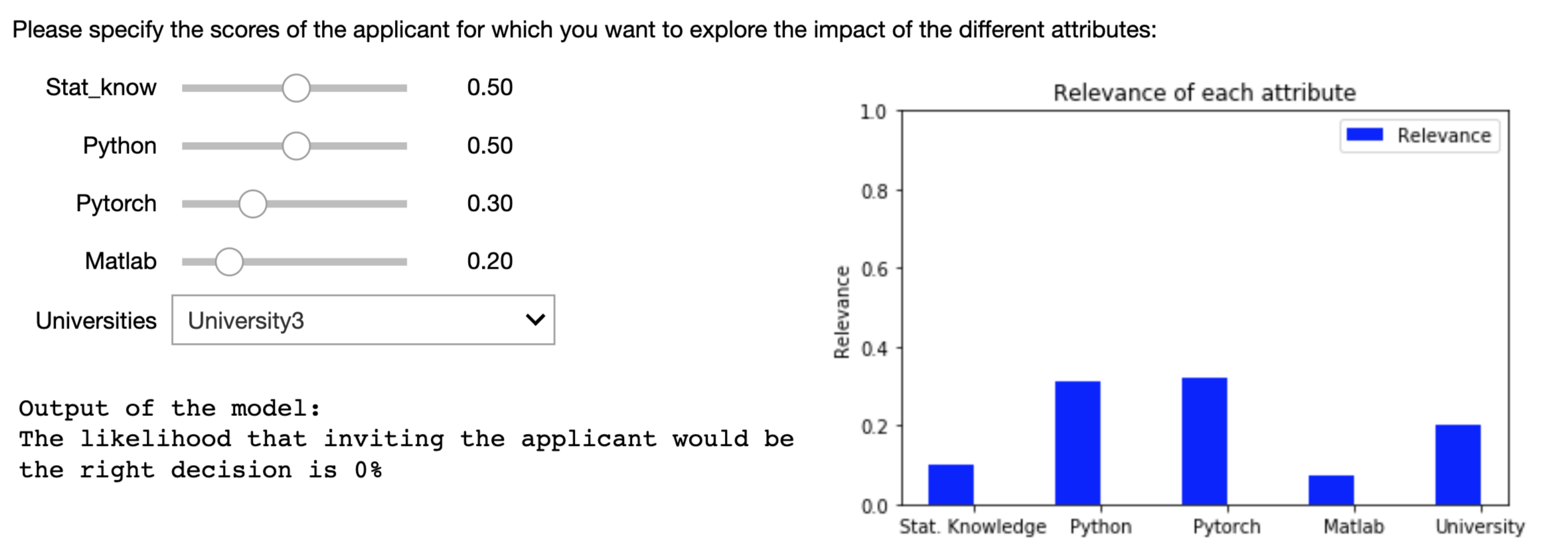}
        \caption{Interactive Visualization of Relevance of each Feature}
        \label{fig:SampleS_inf}
\end{figure}

As for the Covariate Shift we can also look at the relevance that the different features had on a certain decision to further explore the model (see Figure \ref{fig:SampleS_inf}). This time we are most interested in the impact that the university that an applicant comes from has. Remember that when we applied the tool to the data with the Covariate Shift, changing the university did not have an impact at all on the relevance scores. This time the impact is significant! Especially, University10, University9, and University3 are considered as relevant indicators by the neural network for an increased likelihood that the applicant should be invited to an interview. The effect is strongest for University10 for which the examples in the training data had the strongest correlation to our target variable. On the other hand, the fact alone that an applicant graduated from one of these three universities is not enough for the model to recommend to invite the candidate. This shows that the model correctly recognized that high scores are an important factor for the decision as well. Try selecting different universities and scores to see the effect!

\subsection{Imbalance Bias}
An Imbalance Bias occurs if there is a significant difference in the number of samples in the training data for the different labels. In our example scenario this type of bias could occur if the company gets significantly more applications from people that clearly do not fit to the job posting than they get applications from promising candidates. As a consequence many more people would be rejected right away than invited to an interview.

This type of bias is easy to detect. We only have to count how many examples for each label exist in the training data set and check if this significantly differs between the labels. The pie chart below (left) shows the ratio of applicants that were invited to a job interview (blue) in comparison to the percentage of applicants that were rejected in the past (orange). In the next pie chart, we see the same information for today’s applicants for which we want to test our model. By dragging the slider you can change the ratio of invite / reject decisions in the training data, thereby increasing or decreasing the degree of bias.

Let us assume that at the time when we collected the training data about 10\% of the applicants were invited to a job interview. In the test dataset we have a similar ratio of applicants that fit or do not fit the job description. Consequently, the HR department would invite about the same percentage of applicants as the years before. We can simulate such a situation by setting the value of the slider to 10\%.

\begin{figure}[h]
        \centering
        \includegraphics[width=\textwidth]{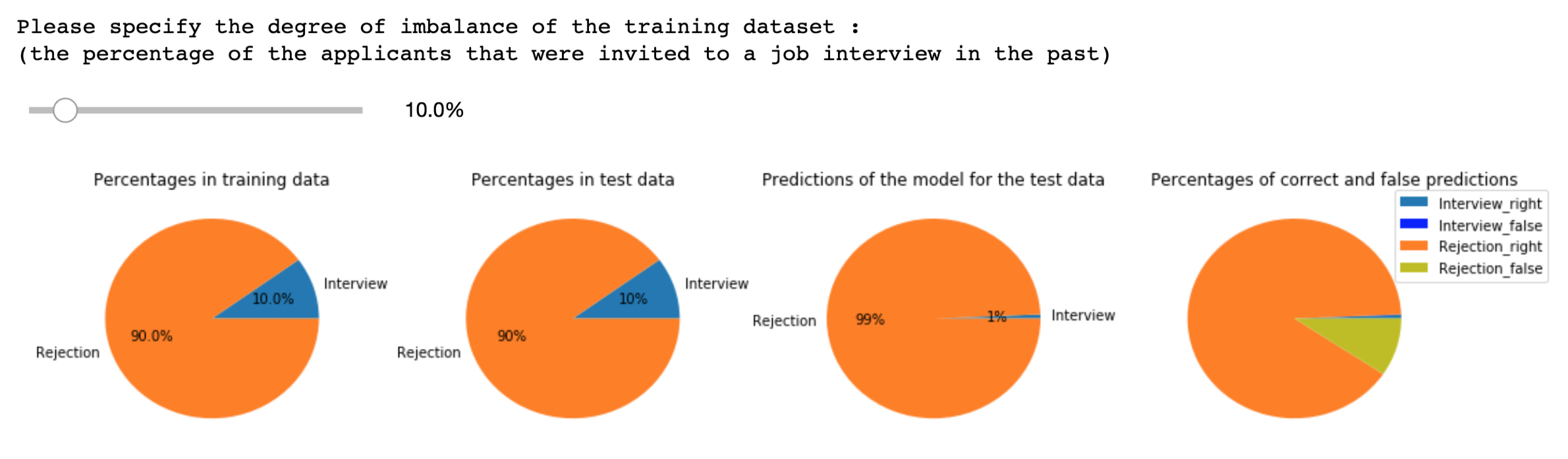}
        \caption{Interactive Visualization of Relevance of each Feature}
        \label{fig:Imb_inf}
\end{figure}

In Figure \ref{fig:Imb_inf}, as you can see in the first two pie charts, now the training dataset and the test dataset have the same percentage of applicants that the HR department would invite to an interview. The third pie chart shows the ratio of invite / reject recommendations of the model when it is applied to the test data. As you can see, the model is biased towards rejection. The last pie chart provides additional details. It further splits up the two categories into right and wrong reject / invite decisions.

If you drag the slider towards the middle, choosing a ratio of about 50\% for the “reject” and “invite” classes in the training data, there is no imbalance bias in the training data anymore. You will then see in the pie charts at the right that the model gets close to the correct ratio of 10 to 90 which have in the test data. However, the last pie chart in the row reveals that the decision was not always correct for the single applicants. (Which is not an issue of bias though, but a sign that our model in general still could be improved.)

Try other values, e.g. a setting in which significantly more applicants were invited than rejected in the training data, to see how this affects the model’s results.

\section{Conclusions}
Is every bias that exists in the training data automatically picked up by a machine learning model? No, this is not necessarily the case. In reality the relationships are much more complex than in the artificial datasets which we used in this article and the bias may not be strong enough to take effect. However, it can happen and this alone commands the need to take countermeasures and inspect a model before it is deployed; especially in applications in which the decision that a machine learning model takes has a significant impact on individuals or our society.

In this article we have employed visualization techniques and methods of Explainable Artificial Intelligence to illustrate that a machine learning model can pick up bias in data. Similar techniques can be used as a means to inspect the training data and the resulting machine learning models.

In the introduction we cited blog posts and articles claiming that machine learning techniques can help to eliminate bias in decisions. So, is this really the case? In this article we have shown that if the data is biased, the resulting machine learning model may be biased, too. However, machine learning experts are able to take countermeasures against this, provided they are aware that a certain type of bias might exist in the data. This necessitates a close collaboration between domain experts and machine learning experts and the willingness to inspect and challenge the data and the resulting model before it is deployed. But if we take the additional effort, machine learning may indeed help us to come our goal of a world without bias a little closer.

\section{Further Reading}
The above article is meant to be a gentle introduction to the topic of bias in machine learning. Its goal is to raise awareness and to increase the understanding of the readers on the different types of bias that can occur in a machine learning model. In the following, we provide additional information on a scientific level for interested readers that want to dig deeper into the topic.

\textbf{Bias} in datasets can lead to unfairness and discrimination such as gender bias, racism \cite{buolamwini2018gender,bolukbasi2016man,zhao2017men}. More generally, bias in datasets also has an impact on the performance of the trained model. Even popular benchmark datasets are often biased, which calls for building a more objective training dataset \cite{torralba2011unbiased,khosla2012undoing,tommasi2017deeper}.

\textbf{Explainability} has received increased attention both in research communities and in the society. A number of attribution methods are proposed. Model-agnostic methodologies explain the decision of the applied model without exploring the inner workings, such as Local Interpretable Model-agnostic Explanations (LIME) \cite{ribeiro2016should}, Partial Dependence Plots \cite{zhao2019causal}, Permutation Variable Importance \cite{altmann2010permutation}. Since the neural networks achieve state-of-the-art performance and extreme lack of transparency, most proposed model-specific methodologies are specific for the neural networks. More concretely, the forward perturbation methods observe the changes of output in case of perturbating the input features \cite{zeiler2014visualizing,zintgraf2017visualizing}. Backpropagation-based methods compute the attributions for all input features in forward and backward passes through the network. The attributions are the vanilla gradients or their modifications \cite{Simonyan2013DeepIC,springenberg2014striving,bach2015pixel,sundararajan2017axiomatic,smilkov2017smoothgrad,selvaraju2017grad,ribeiro2016should,shrikumar2017learning,Gu2018UnderstandingID}.

\bibliography{vis}
\bibliographystyle{unsrt}

\end{document}